\documentclass[runningheads]{llncs}

\usepackage{eccv}

\usepackage{eccvabbrv}

\usepackage{graphicx}
\usepackage{booktabs}
\usepackage{bbding}
\usepackage{makecell}
\usepackage{amsfonts}
\usepackage{amssymb}
\usepackage{setspace} 
\usepackage{caption}

\usepackage{marvosym}
\usepackage[accsupp]{axessibility}  
\usepackage{wrapfig}
\usepackage{xurl}

\usepackage{hyperref}

\usepackage{orcidlink}
\makeatletter
\def\thanks#1{\protected@xdef\@thanks{\@thanks
        \protect\footnotetext{#1}}}

\begin{document}

\title{RangeLDM: Fast Realistic LiDAR Point Cloud Generation} 


\author{Qianjiang Hu\orcidlink{0009-0002-9375-1008} \and
Zhimin Zhang\orcidlink{0000-0001-7148-1129}\and
Wei Hu\textsuperscript{\Letter}\orcidlink{0000-0002-9860-0922}
\thanks{
\textsuperscript{\Letter} Corresponding author: W. Hu.
}
}

\authorrunning{Q.~Hu et al.}

\institute{Wangxuan Institute of Computer Technology, Peking University, Beijing, China\\
\email{hqjpku@pku.edu.cn}
\email{zm\_zhang@stu.pku.edu.cn}
\email{forhuwei@pku.edu.cn}}

\maketitle
\begin{center}
    \centering
    \includegraphics[width=\textwidth]{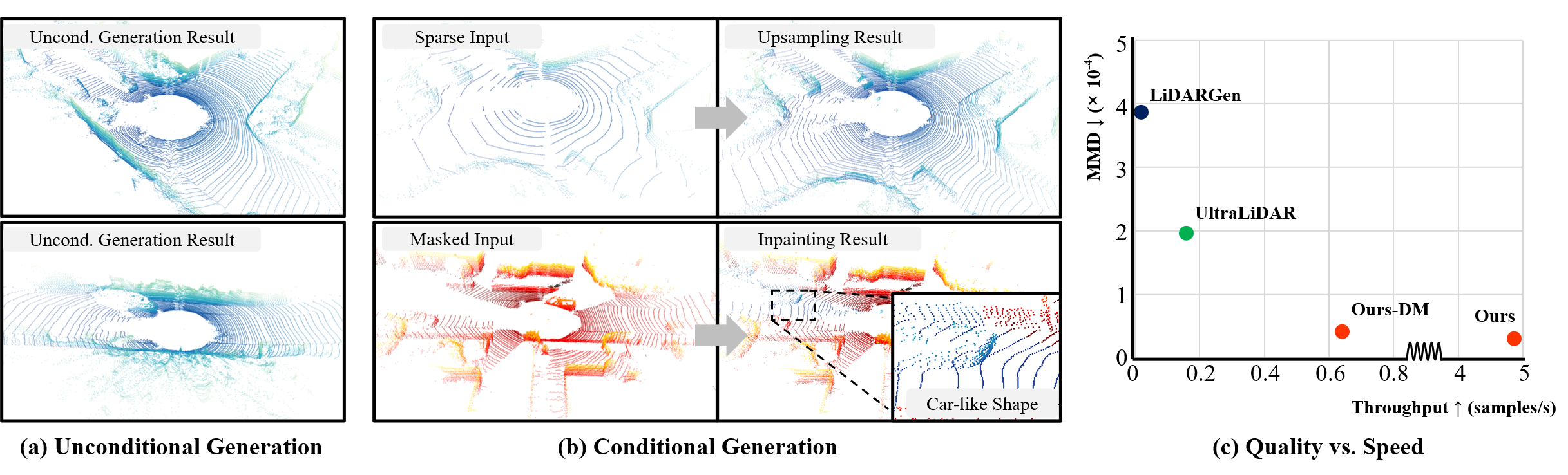}
    \captionof{figure}{\textbf{(a).} Unconditional LiDAR point cloud generation with realistic global structure. \textbf{(b).} Conditional LiDAR point cloud generation, including LiDAR point cloud upsampling and inpainting. \textbf{(c).} Generation quality (Maximum Mean Discrepancy, \textit{abbr.} MMD) vs. generation speed (samples/s) of competitive LiDAR point cloud generation methods on the KITTI-360~\cite{liao2022kitti} dataset. The proposed method outperforms the state-of-the-art methods LiDARGen \cite{zyrianov2022learning} and UltraLiDAR \cite{xiong2023learning} in both generation quality and generation speed. All speeds are evaluated on a single RTX 3090 GPU. }
    \label{fig:teaser}
\end{center}

 \begin{abstract}
Autonomous driving demands high-quality LiDAR data, yet the cost of physical LiDAR sensors presents a significant scaling-up challenge. 
While recent efforts have explored deep generative models to address this issue, they often consume substantial computational resources with slow generation speeds while suffering from a lack of realism. 
To address these limitations, we introduce RangeLDM, a novel approach for rapidly generating high-quality range-view LiDAR point clouds via latent diffusion models. 
We achieve this by correcting range-view data distribution for accurate projection from point clouds to range images via Hough voting, which has a critical impact on generative learning. 
We then compress the range images into a latent space with a variational autoencoder, and leverage a diffusion model to enhance expressivity. 
Additionally, we instruct the model to preserve 3D structural fidelity by devising a range-guided discriminator.
Experimental results on KITTI-360 and nuScenes datasets demonstrate both the robust expressiveness and fast speed of our LiDAR point cloud generation.
\end{abstract}    
\section{Introduction}
\label{sec:intro}

Autonomous systems have drawn great attention in both the academia and the industry community. 
Numerous autonomous systems leverage the power of various sensors and deep learning to enhance the perception of the 3D world. 
LiDAR (\textbf{Li}ght \textbf{D}etection \textbf{A}nd \textbf{R}anging), with its ability to provide precise 3D geometric information about the surroundings, has become a popular sensor choice for autonomous systems, including self-driving cars~\cite{shi2019pointrcnn, yin2021center, shi2020pv}, surveying drones~\cite{zhang2014loam, resop2019drone} and robots~\cite{malavazi2018lidar, weiss2011plant}. 

However, while LiDAR offers accurate geometric measurements, it comes with a significant limitation: the data collection process is exceedingly expensive and challenging to scale up. 
This renders physical sensors impractical for scalable and customizable data collection.
Additionally, it is difficult to collect data in corner cases such as car accidents and extreme weather conditions in the field.

One approach to mitigate this issue is to employ existing LiDAR simulation toolkits~\cite{manivasagam2020lidarsim,li2023pcgen} to synthesize more data.
Nevertheless, these systems typically demand manual scene creation or rely on multiple prior scans of the real world.  
Another approach is leveraging deep generative models to generate LiDAR point clouds. 
This is challenging due to the high degree of unstructuredness, sparsity, and non-uniformity in LiDAR point clouds. 
Caccia et al.~\cite{caccia2019deep} proposed LiDAR GAN and LiDAR VAE to generate LiDAR point clouds with generative adversarial networks (GANs) \cite{goodfellow2020generative} and variational autoencoders (VAEs) \cite{kingma2013auto}, respectively. 
However, the generation results often suffer from issues such as fuzzy or missing details.
LiDARGen \cite{zyrianov2022learning} introduced a novel score-based model to synthesis LiDAR point clouds, which however sampled slowly and failed to generate high-quality geometric details at a far range.
UltraLiDAR~\cite{xiong2023learning} proposed to synthesize voxelized point clouds in bird's-eyes-view (BEV) with vector quantized VAE (VQ-VAE)~\cite{van2017neural}, resulting in more realistic point clouds than previous methods. 
Nevertheless, due to the large size and sparsity of point clouds in BEV, most computing power of UltraLiDAR is devoted to generating empty voxels, which results in a low generation speed.

To this end, we propose RangeLDM, a novel approach based on latent diffusion models (LDMs)~\cite{rombach2022high}, which improves both the quality and speed of LiDAR point cloud generation significantly. 
Firstly, we introduce a range image view to represent point clouds. 
The reasons are twofold: 
1) Range images are compact and closely mimic the sampling conditions of LiDAR, making them a suitable representation.
A LiDAR point cloud essentially constructs a 2.5D scene from a single viewpoint instead of a full 3D point cloud~\cite{hu2020you}. 
Consequently, organizing the point cloud in range view ensures that no information is overlooked.
2) 2D image generation techniques are relatively mature, providing a solid foundation for our work. 
While LiDARGen~\cite{zyrianov2022learning} also represents point clouds as range images, it suffers from blurriness caused by inaccurate projection. 
In contrast, we provide insights that {\it the correct range-view data distribution} has a critical impact on such range-view-based generative model learning, and accurately project point clouds onto range images using parameters estimated by Hough Voting. 
Then, to achieve a faster sampling speed~\cite{rombach2022high} and greater expressivity~\cite{vahdat2021score, rombach2022high}, we compress range images to the latent space with a VAE and develop a diffusion model operating on the latent space, leveraging the successful paradigm of LDMs~\cite{rombach2022high, vahdat2021score, zeng2022lion}. 
Further, in order to enhance the ability of the VAE in reconstructing 3D structures, we introduce a range-guided discriminator, which is supervised from the spherical coordinates and thus geometry-sensitive. 
This plays a crucial role in guiding the decoder to generate high-quality range images, ultimately preserving the fidelity of the 3D structure. 
Extensive experimental results show that the proposed RangeLDM achieves the state-of-the-art performance on KITTI-360~\cite{liao2022kitti} and nuScenes~\cite{caesar2020nuscenes} datasets in terms of both the generation quality and generation speed, as demonstrated in Figure~\ref{fig:teaser} (c). 
The LiDAR upsampling and LiDAR inpainting results also demonstrate the potential of the proposed method for conditional generation, as presented in Figure~\ref{fig:teaser} (a) and (b).

To summarize, the main contributions of this paper include: 
\begin{itemize}
    \item We propose a latent diffusion model to capture the distribution of range-view LiDAR point clouds, aiming to generate realistic point cloud scenes at a fast speed.
    \item We enlighten the significance of the correct range-view data distribution for range-view-based generative models and achieve high-quality range image projection via Hough Voting. 
    \item We exploit a range-guided discriminator to ensure preserving the fidelity of the 3D geometric structure in generated point clouds.
    \item Experimental results on the KITTI-360 and nuScenes datasets highlight that our approach outperforms state-of-the-art methods in terms of both visual quality and especially generation speed.
\end{itemize}

\section{Related Work}
\label{sec:related}

\noindent\textbf{LiDAR Representation.}
LiDAR point clouds can be represented in various forms, encompassing raw point clouds, voxels, range views, and multi-view fusion.
\textbf{Point-based models}~\cite{shi2019pointrcnn, yang2019std, yang20203dssd, shi2020point, pan20213d, wang2022rbgnet, zhang2022not} directly encode 3D objects from raw points using the PointNet~\cite{qi2017pointnet} encoder and subsequently perform detection or segmentation based on point features.
These methods wholly preserve the irregularity and locality of a point cloud but have relatively higher latency.
\textbf{Voxel-based methods}~\cite{engelcke2017vote3deep, li20173d, yang2018pixor, zhou2018voxelnet, yan2018second, lang2019pointpillars, wang2020pillar, shi2020points, deng2021voxel, yin2021center, shi2022pillarnet, sun2022swformer, fan2022embracing} voxelize point clouds for convolutional neural networks to efficiently capture features. 
They are computationally effective but the desertion of fine-grained patterns degrades further refinement.
Given that the range view is compact and compatible with the LiDAR sensor's sampling process, \textbf{range-image-based methods}~\cite{fan2021rangedet, sun2021rsn, chai2021point, chen2021range, li2016vehicle, meyer2019lasernet, tian2022fully, milioto2019rangenet++, xu2020squeezesegv3, cortinhal2020salsanext, kochanov2020kprnet, zhao2021fidnet, ando2023rangevit, cheng2022cenet} have been proposed to directly process range images for LiDAR perception. 
\textbf{Multi-view fusion methods}~\cite{shi2020pv, he2022voxel, hu2022point} amalgamate multiple representations, which yield better results at the expense of processing speed.
In this paper, we adopt the range-view representation, which is congruent with the LiDAR sensor's sampling process and can be efficiently encoded by image generation models.

\noindent\textbf{Generative Models for Point Clouds.}
Given observed samples of interest, generative models aim to learn the underlying distribution of the data and generate new samples.
The first 3D generative models~\cite{achlioptas2018learning, shu20193d, valsesia2018learning} for point clouds are based on GANs~\cite{goodfellow2020generative}. 
They operate by generating point clouds and discriminating them from real samples in an adversarial manner.
The second category of methods~\cite{litany2018deformable, tan2018variational, mo2019structurenet, gao2019sdm, gao2021tm, kim2021setvae, mittal2022autosdf} employ VAEs~\cite{kingma2013auto} to explore the probabilistic latent space of 3D shapes. 
Auto-regressive models~\cite{box2015time} have also been introduced to generate point clouds~\cite{sun2020pointgrow} from scratch or semantic contexts.
PointFlow~\cite{yang2019pointflow} and SoftFlow~\cite{kim2020softflow} leverage normalizing flows~\cite{rezende2015variational} to capture the likelihood of shapes and points.
Notably, the recent success of denoising diffusion models (DDMs)~\cite{ho2020denoising} for image synthesis~\cite{rombach2022high} has also been extended to the domain of 3D point cloud generation~\cite{luo2021diffusion, zhou20213d, cai2020learning, zeng2022lion, nichol2022point, hui2022neural}. 
However, these previous works often concentrate on the object level and are not well-suited for handling large scenes.

In this paper, our focus lies in the generation of LiDAR point clouds. 
While~\cite{caccia2019deep} and~\cite{sallab2019lidar} employ VAE or GANs for LiDAR point cloud generation, the realism achieved in their results is relatively limited. 
UltraLiDAR~\cite{xiong2023learning}, on the other hand, utilizes VQ-VAE to generate voxelized LiDAR point clouds, but at the expense of introducing quantified losses and slow generation speed. 
LiDARGen~\cite{zyrianov2022learning} proposed a novel score-matching energy-based model to generate higher-quality LiDAR point clouds, but it samples very slowly and has degraded geometric details at a far range.
A contemporary work LiDM~\cite{ran2024towards} also utilized LDMs to generate LiDAR point clouds. 
However, while they focused on generating LiDAR point clouds conditional on multimodal data, we focus on improving the quality of point cloud generation directly. 
\section{Background on Denoising Diffusion Models}
\label{sec:back}
DDMs~\cite{ho2020denoising} are latent variable models that employ a pre-defined posterior distribution, known as the forward diffusion process, and are trained with a denoising objective. 
More specifically, given samples $\mathbf{x}_0 \sim q\left(\mathbf{x}_0\right)$ from a data distribution, DDMs follow the form $p_\theta\left(\mathbf{x}_0\right):=\int p_\theta\left(\mathbf{x}_{0: T}\right) d \mathbf{x}_{1: T}$, where $T$ denotes the number of steps, 
$\mathbf{x}_1, ..., \mathbf{x}_T$ are latent variables that gradually add noise to the data $\mathbf{x}_0$, 
and $\theta$ denotes the parameter set of the DDM decoder.

DDMs are trained by minimizing the evidence lower bound (ELBO) of the data $\mathbf{x}_0$ under $p_{\boldsymbol{\theta}}\left(\mathbf{x}_{0: T}\right)$. 
This objective can be simplified to~\cite{ho2020denoising}:
\begin{equation}
\min_\theta \mathbb{E}_{\mathbf{x}, \boldsymbol{\epsilon} \sim \mathcal{N}(0,1), t}\left[\left\|\boldsymbol{\epsilon}-\boldsymbol{\epsilon}_\theta\left(\mathbf{x}_t+\boldsymbol{\epsilon}, t\right)\right\|_2^2\right], 
\label{eq:train_ddm}
\end{equation}
with $t$ uniformly sampled from $\{1, ..., T \}$.

\section{The Proposed Method}
\label{sec:method}
\begin{figure*}[t]
    \centering
    \includegraphics[width=\textwidth]{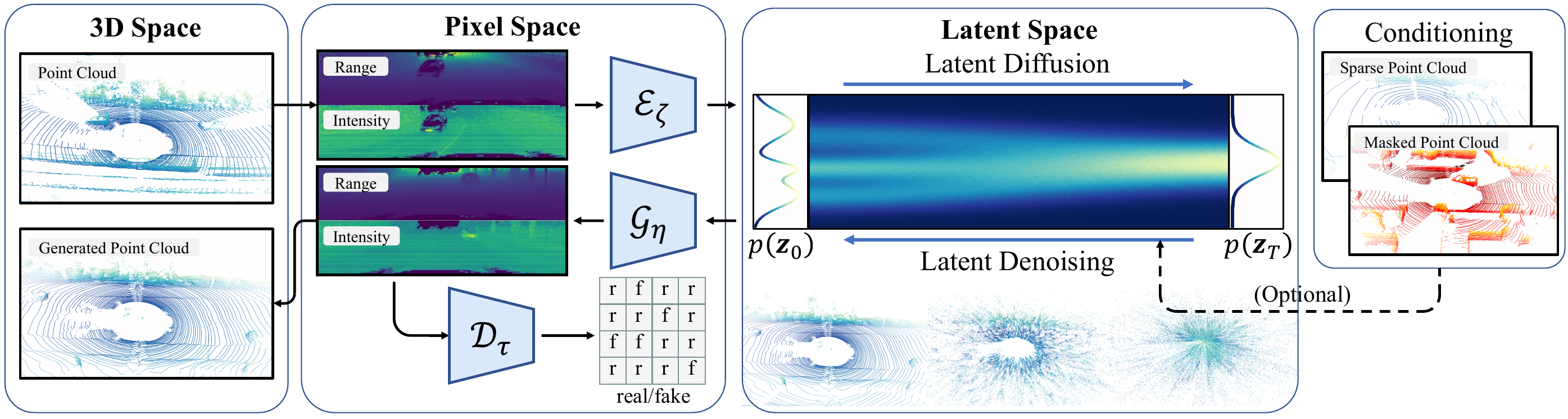}
    \caption{The framework of the proposed RangeLDM. Firstly, we project point clouds onto high-quality range images via Hough Voting (Section~\ref{subsec:range_projection}). Subsequently, we train a VAE to compress the range images into low-dimensional latent features $\mathbf{z}_0$, which encodes the range images with the encoder $\mathcal{E}_\zeta$ and reconstructs range images from latent features with the decoder $\mathcal{G}_\eta$ (Section~\ref{subsubsec:vae}). Here, a range-guided discriminator $\mathcal{D}_\tau$ is introduced to guide the decoder in the reconstruction of 3D structures. We finally train a latent diffusion model to  capture the distribution of the latent features (Section~\ref{subsubsec:diffusion}). With optional conditional inputs, the proposed method is applicable to tasks such as point cloud upsampling and inpainting (Section~\ref{sec:generation_tasks}).}
    \label{fig:framework}
\end{figure*}

As shown in Fig.~\ref{fig:framework}, we first project point clouds onto high-quality range images (as detailed in Section~\ref{subsec:range_projection}), considering the compact nature of the range view, which aligns seamlessly with the sampling process of LiDAR sensors. 
Our model training process is then focused on these projected 2D range images.
We then compress range images into a low-dimensional latent space through a VAE (as detailed in Section~\ref{subsubsec:vae}).  
Subsequently, we proceed to train DDMs within the reduced-dimensional latent space (as discussed in Section~\ref{subsubsec:diffusion}).
We discuss applications of the proposed model in Section~\ref{sec:generation_tasks}, which involves unconditional generation and conditional generation. 
Implementation details are explained in Section~\ref{subsec:implementation_details}.

\subsection{High-Quality Range Projection}
\label{subsec:range_projection}

The range image presents LiDAR data compactly and intuitively, with rows indicating the laser beams and columns representing the yaw angles.
We convert point clouds to range images using spherical projection. Typically, for a point $\mathbf{p}$ in Cartesian coordinates $(x,y,z)$, we calculate its spherical coordinates $(r,\theta,\phi)$ using:

\begin{equation}
\label{eq:ori_sph}
r=\sqrt{x^2+y^2+z^2},  
\theta=\operatorname{atan}(y, x), 
\phi=\operatorname{atan}\left(z, \sqrt{x^2+y^2}\right).
\end{equation}

\begin{figure}[t]
\parbox{.55\linewidth}{
\centering
\includegraphics[width=\linewidth]{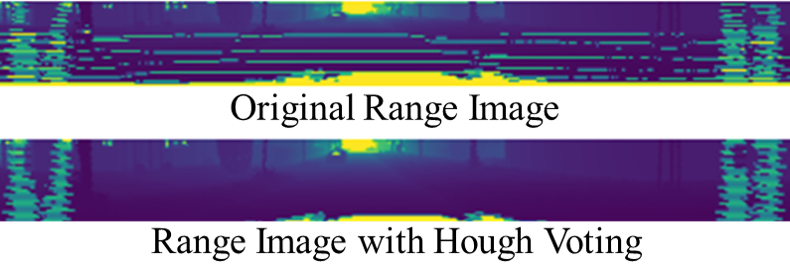}
\caption{Comparison of range projection by the typical method described in Eq.~\ref{eq:ori_sph} and our method with Hough Voting as in Eq.~\ref{eq:new_spi}. }
\label{fig:range_image}
}
\hfill
\parbox{.42\linewidth}{
\centering
\centering
\resizebox{\linewidth}{!}{
\begin{tabular}{l|ccc}
\hline
Method &  $\text{MMD}_\text{BEV}$ $\downarrow$ & $\text{FRD}$ $\downarrow$ & $\text{JSD}_\text{BEV}$ $\downarrow$ \\
\hline
LiDARGen & $3.87 \times 10^{-4}$ & 2040.1 & 0.067 \\
~~~+Hough Voting & $\mathbf{1.41 \times 10^{-4}}$ & $\mathbf{1453.1}$ & $\mathbf{0.064}$ \\
\hline
\end{tabular}
}
\captionof{table}{Unconditional generation performance of LiDARGen~\cite{zyrianov2022learning} with and without Hough Voting.
Hough Voting provides performance improvements through precise range image projection.}
\label{tab:lidargen}
}
\end{figure}

However, in most current datasets such as KITTI-360, multiple lasers from the Velodyne LiDAR system do not share a common origin for their measurements. This may introduce errors in the direct conversion from Cartesian points to spherical points, resulting in incorrect range-view data distribution and thus low-quality range images, as shown at the top of Fig.~\ref{fig:range_image}.

To address this issue, we adopt Hough Voting to estimate heights and pitch angles $\{h_j, \phi_j\}_{j=1,...,N}$ for Velodyne sensors~\cite{RCD}. 
We then adjust the point cloud transformation to a range image using 
\begin{equation}
\label{eq:new_spi}
r=\sqrt{x^2+y^2+(z-h_j)^2},  
\theta=\operatorname{atan}(y, x), 
\phi=\phi_j, 
\end{equation}
where $h_j$ and $\phi_j$ refer to the $j$-th Velodyne sensor.

We then rasterize points $(r, \theta, \phi)$ into a 2D cylindrical projection $R(u, v)$ (a.k.a., range image) of size $H \times W$ with $u=\left( \left( \theta +\pi \right) /2\pi \right) W, v=j,$
where $(u, v)$ denotes the grid coordinate of a point in the range image.
Thus, we obtain high-quality range images as illustrated at the bottom of Fig.~\ref{fig:range_image}. 
We denote the obtained range image as $\mathbf{x} \in \mathbb{R}^{H \times W \times 2}$, which comprises $H \times W$ pixels associated with both range and intensity $\{r, i\}$. 

To verify the impact of range projection, we trained range-based LiDAR generation methods like LiDARGen~\cite{zyrianov2022learning} with our obtained high-quality range images. 
Table~\ref{tab:lidargen} shows substantial improvement, which sheds light on the significance of the correct range-view data distribution for such range-based generative model learning.

\subsection{Training}
\label{subsec:network_architecture}

As a range image is often of high dimensionality at the scale of $64 \times 1024$, it would be computationally expensive to learn the distribution directly. 
Instead, we compress the range image into a low-dimensional latent space via a VAE and then model the distribution via an LDM.     
Consequently, the training of the proposed RangeLDM includes two distinct stages.
In the first stage, we train a regular VAE to compress the range image into latent features $\mathbf{z}_0 \in \mathbb{R}^{h \times w \times c}$ by a downsampling factor $f=H/h=W/w$.
In the second stage, we train an LDM to learn the distribution of the latent encoding $\mathbf{z}_0$.

\noindent\textbf{First-Stage: Dimensionality Reduction.}
\label{subsubsec:vae}
In the first stage, we reduce the dimensionality of the obtained range images by a VAE, which removes imperceptible high-frequency details and leads to low-dimensional latent features that encode prominent information in the original range image.   

A standard VAE consists of two main components: 1) an encoder $\mathcal{E}_{\zeta}$ that transforms an input range image $\mathbf{x} \in \mathbb{R}^{H \times W \times 2}$ into latent features $\mathbf{z}_0 = \mathcal{E}_{\zeta}(\mathbf{x}) \in \mathbb{R}^{h \times w \times c}$ and 2) a decoder $\mathcal{G}_{\eta}$ that takes the latent features $\mathbf{z}_0$ as input and generates the reconstructed range image $\hat{\mathbf{x}} = \mathcal{G}_{\eta}(\mathbf{z}_0)$.
The entire VAE is trained by maximizing a modified ELBO with respect to the parameters $\zeta$ and $\eta$~\cite{kingma2013auto, rezende2014stochastic}:
\begin{equation}
\mathcal{L}_{ELBO}(\zeta, \eta)= \mathbb{E}_{q_\zeta(\mathbf{z}_0\mid\mathbf{x})}\left[\log p_{\boldsymbol{\eta}}(\mathbf{x} \mid \mathbf{z}_0)\right]  -\lambda_1 D_{\mathrm{KL}}\left(q_{\boldsymbol{\zeta}}(\mathbf{z}_0 \mid \mathbf{x}) \| p(\mathbf{z}_0)\right).
\end{equation}
In the above equation,  the first term evaluates the reconstruction likelihood of the decoder from the latent $\mathbf{z}_0$ and corresponds to an $L_1$ reconstruction loss, while the second term quantifies how closely the learned distribution of $\mathbf{z}_0$ resembles the prior distribution $p(\mathbf{z}_0)=\mathcal{N}(\mathbf{z}_0; \mathbf{0}, \mathbf{I})$ held over latent variables. 
The hyperparameter $\lambda_1$ balances the reconstruction accuracy and Kullback-Leibler regularization.

\noindent\textbf{Range-Guided Discriminator.}
In order to mitigate the blurriness introduced by relying solely on the pixel-space $L_1$ reconstruction loss, we integrate the VAE with a patch-based adversarial discriminator~\cite{isola2017image, esser2021taming, yu2021vector, rombach2022high}. 
Furthermore, to ensure that the reconstruction remains confined within the manifold underlying the range image and exploit the geometric information from the spherical coordinates, we propose a range-guided discriminator $\mathcal{D}_\tau$, with parameters $\tau$. 
Specifically, we adapt the Meta-Kernel~\cite{fan2021rangedet} to replace the standard convolution in the discriminator, aiming to learn convolution weights from relative spherical coordinates. 
The Meta-Kernel is formulated as
\begin{equation}
\mathbf{h}_i'=\mathcal{W}\left(\mathop{\mathcal{A}}\limits_{j \in \mathcal{N}(i)} \left(\Phi\left(\gamma(\mathbf{p}_j, \mathbf{p}_i)\right) \odot \mathbf{h}_j\right)\right),
\end{equation}
where $\mathbf{h}$ and $\mathbf{h}'$ represent the input feature and output feature, respectively. 
The function $\Phi$ denotes a Multi-Layer Perceptron (MLP) with two fully-connected layers, $\mathcal{A}$ is a concatenation operation, $\mathcal{N}(i)$ denotes the 2D-grid neighborhood around the center pixel $i$, and $\mathcal{W}$ is a fully-connected layer. 
The term $\gamma(\mathbf{p}_j, \mathbf{p}_i)$ represents the distance between points $\mathbf{p}_j$ and $\mathbf{p}_i$ in spherical coordinates, which is expressed as
\begin{equation}
\begin{gathered}
\gamma\left(\mathbf{p}_j, \mathbf{p}_i\right):=\{r_j \cos (\Delta \theta) \cos (\Delta \phi)-r_i, r_j \cos (\Delta \theta) \sin (\Delta \phi),  r_j \sin (\Delta \theta)\}, \\
\Delta \theta=\theta_j-\theta_i, \quad \Delta \phi=\phi_j-\phi_i.    
\end{gathered}
\end{equation}
The Meta-Kernel is aware of local 3D structures by adaptively adjusting convolution kernel weights from relative spherical coordinates. 
This makes it challenging for the decoder to deceive the range-guided discriminator. 
This, in turn, encourages the decoder to generate more realistic range images.

With the range-guided discriminator, the final objective of the first-stage training is:
\begin{equation}
\zeta, \eta= \arg\mathop{\min}\limits_{\zeta, \eta} \mathop{\max}\limits_{\tau} \mathcal{L}_{ELBO}(\zeta, \eta) + \lambda_2 \mathcal{L}_{adv}(\zeta, \eta, \tau),
\end{equation}
where $\mathcal{L}_{adv}$ is the GAN Hinge loss~\cite{lim2017geometric} and $\lambda_2$ balances the two losses.

\noindent\textbf{Second-Stage: Latent Diffusion Modeling}
\label{subsubsec:diffusion}
In the second stage, we freeze the encoder and decoder of the learned VAE and train an LDM on the encoded latent feature $\mathbf{z}_0$ for distribution modeling of the range image.
During the diffusion process, noise is added to the initial latent $\mathbf{z}_0$, resulting in a noisy latent 
$\mathbf{z}_t$,  
where the noise level increases over time steps $t \in T$.

Regarding the modeling of the unconditional distribution $p(\mathbf{z}_0)$, based on Eq.~\ref{eq:train_ddm}, we learn a time-conditional UNet \cite{ronneberger2015u} $\boldsymbol{\epsilon}_\theta\left(z_t, t\right)$ that
predicts the noise added to the noisy latent $\mathbf{z}_t$:
\begin{equation}
L_\text{LDM}:=\mathbb{E}_{\mathcal{E}(x), \boldsymbol{\epsilon} \sim \mathcal{N}(0,1), t}\left[\left\|\boldsymbol{\epsilon}-\boldsymbol{\epsilon}_\theta\left(z_t, t\right)\right\|_2^2\right].
\end{equation}

As to the modeling of the conditional distribution $p(\mathbf{z}_0 | \mathbf{y})$, where $\mathbf{y}$ is a condition such as sparse or masked point clouds, we learn a conditional LDM via
\begin{equation}
L_\text{LDM}:=\mathbb{E}_{\mathcal{E}(x), \mathbf{y}, \boldsymbol{\epsilon} \sim \mathcal{N}(0,1), t}\left[\left\|\boldsymbol{\epsilon}-\boldsymbol{\epsilon}_\theta\left(\mathbf{z}_t, t, \tau_\theta(\mathbf{y})\right)\right\|_2^2\right],
\label{eq:conditional_training}
\end{equation}
where $\tau_\theta$ is a condition encoder that projects $\mathbf{y}$ into an intermediate representation $\tau_\theta(\mathbf{y})$,  and $\epsilon_\theta\left(\mathbf{z}_t, t, \tau_\theta(\mathbf{y})\right)$ is a UNet interpolated with cross-attention layers~\cite{vaswani2017attention,rombach2022high}.

\subsection{Generation}
\label{sec:generation_tasks}
\noindent\textbf{Unconditional Generation.}
With the decoder $\mathcal{G}_{\eta}$ and the LDM $\epsilon_\theta$, we establish a hierarchical generative model $p_{\eta,\theta}(\mathbf{x}, \mathbf{z}_0) = p_\eta(\mathbf{x}|\mathbf{z}_0)p_\theta(\mathbf{z}_0)$. 
That is, we first generate latent features with the LDM, and then map the latent features back to the original range image space using the decoder $\mathcal{D}_{\eta}$ for LiDAR point cloud generation.

\noindent\textbf{Conditional Generation.}
Given a condition $\mathbf{y}$, the conditional generative model is defined as $p_{\eta,\theta}(\mathbf{x}, \mathbf{z}_0 | \mathbf{y}) = p_\eta(\mathbf{x}|\mathbf{z}_0, \mathbf{y})p_\theta(\mathbf{z}_0 | \mathbf{y})$.  
There could be various conditions for LiDAR point cloud generation, such as sparse point clouds, incomplete point clouds, camera or even text.
Here, we illustrate two applications of conditional generation: LiDAR point cloud upsampling and inpainting.

\textit{ LiDAR Point Cloud Upsampling.}
This task takes a sparse point cloud\footnote{Without causing ambiguity, the point clouds are converted to range images by default.} $\mathbf{y} \in \mathbb{R}^{h \times W \times 2}$ as input (following the setting of LiDARGen) and expects the model to produce a denser counterpart $\hat{\mathbf{x}} \in \mathbb{R}^{H \times W \times 2}$, which is necessary for accurate LiDAR-based perception and density-insensitive domain adaptation.
We train LDM following Eq.~\ref{eq:conditional_training} with the sparse point cloud $\mathbf{y}$ and its ground-truth dense variant $\mathbf{x}$. 
The condition encoder $\tau_\theta$ is configured as a reshaper that transforms the sparse point cloud $\mathbf{y}$ into 
$\mathbf{y}' \in \mathbb{R} ^ {h \times w \times 2f}$.
The reshaped $\mathbf{y}'$ is then concatenated with noisy latent variable $\mathbf{z}_t$ and fed into the UNet $\epsilon_\theta$ for noise prediction. 

\textit{ LiDAR Point Cloud Inpainting.}
This task is required when LiDAR sensors cannot capture the entire scene due to object occlusion or sensor limitations.
We train an inpainting LDM from the ground truth point cloud $\mathbf{x}$, a masked point cloud $\mathbf{x}'$, and the corresponding mask $\mathbf{m} \in \{0, 1\}^{H \times W}$. 
The masked point cloud $\mathbf{x}'$ is encoded with the encoder $\mathcal{E}_\zeta$ from the VAE and is then concatenated with a downsampled mask $\mathbf{m}' \in \{0, 1\} ^ {h \times w}$ as the condition term in Eq.~\ref{eq:conditional_training}. 
Mathematically, we represent the condition $\mathbf{y} $ as $ \{\mathbf{x}', \mathbf{m}\}$, and deploy the condition encoder to project it into $\tau_\theta(\mathbf{y}) = \mathcal{E}_\zeta(\mathbf{x}') \| ds(\mathbf{m}) $, where $\|$ denotes the concatenation operation and $ds(\cdot)$ indicates the downsampling operation. 
The projected condition is then concatenated with the noisy latent variable $\mathbf{z}_t$ and input into the UNet $\epsilon_\theta$ for noise prediction. 

\subsection{Implementation Details}
\label{subsec:implementation_details}

\noindent\textbf{Circular Convolution.}
Range images are inherently circular, which means that the left boundary of a range image is actually connected to its right boundary. 
However, standard convolutions employ zero-padding or symmetry padding and do not take into account such constraints. 
Thus, as in LiDARGen~\cite{zyrianov2022learning}, we replace standard convolutions in both the VAE and LDM with circular convolutions~\cite{schubert2019circular}, which treat the left and right boundaries as connected neighbors in the topology. 
Since the 2D convolution operator satisfies translation invariance, it is evident that the entire network is invariant to horizontal shifting of the range image (\textit{i.e.}, rotation in the xy-plane of the point cloud).

\begin{figure*}[t]
    \centering
    \includegraphics[width=\textwidth]{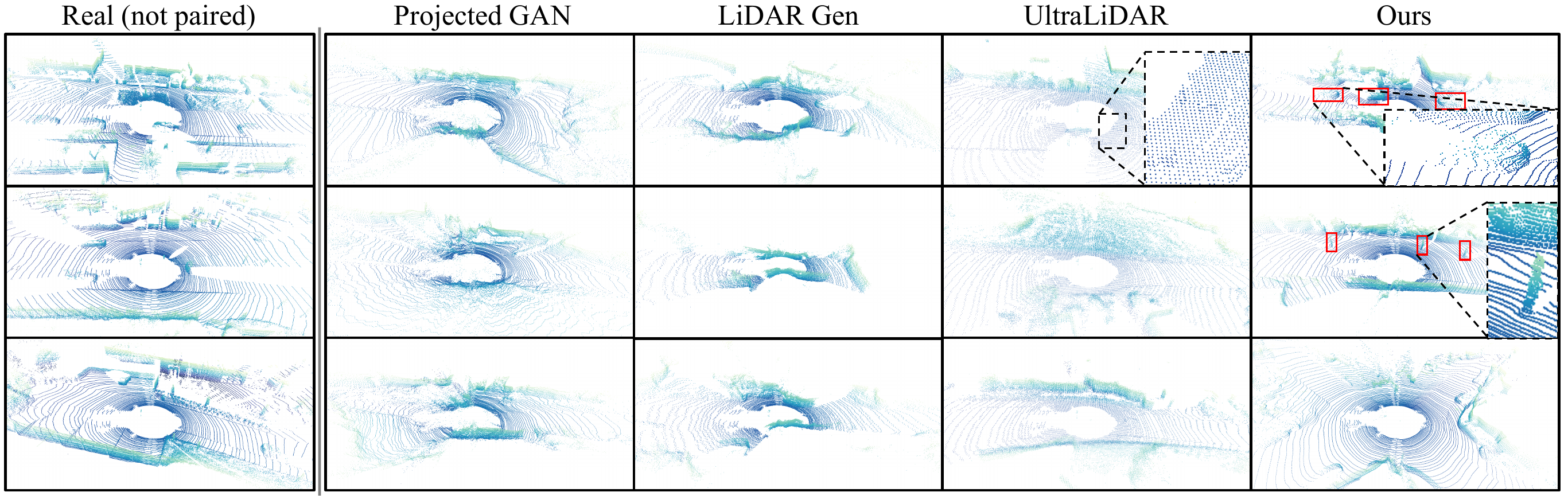}
    \caption{\textbf{Qualitative results comparing against baselines for unconditional LiDAR generation on KITTI-360}. Real point clouds are only for reference. Our model produces results that closely resemble real-world data, which excels in generating road scenes, such as cars (the first row), road bollards (the second row) and crossroads (the last row).}
    \label{fig:unconditional_kitti360}
\end{figure*}

\noindent\textbf{Conditioning the Model with Direction.}
Real LiDAR point clouds collected from driving scenes usually have a certain directionality. 
For example, since vehicles typically follow the direction of the road, the x-axis direction of the LiDAR point cloud usually aligns with the road direction. 
However, since the whole network of our model is {\it invariant} to rotation in the xy-plane of the point cloud, the point clouds generated from random noise could result in arbitrary directions. 
To avoid this, we generate point clouds with directional conditioning. 
Specifically, we use an $h \times w$ matrix as the condition, in which only the values from the first column are set to $1$, while the rest are all set to $0$. 
This $\mathbf{1}$-valued column corresponds to the x-axis of the point cloud. 
Such simple conditioning demonstrates effective control over the direction of the generated point cloud, as presented in the qualitative results of the supplementary material.

\noindent\textbf{Network Architectures and Model Hyperparameters.}
For simplicity, we adopted a similar architecture to that in \cite{rombach2022high}. 
The downsampling factor $f$ is set as $4$ to strike a good balance between efficiency and realistic results. 
The VAE comprises three encoder blocks and three decoder blocks.
The LDM consists of four downsampling blocks with the last three followed by a transformer layer, and four upsampling blocks with the first three blocks also followed by a transformer layer. 
In the generation process, we employed 50 denoising steps with DDIM sampler~\cite{song2020denoising} for point cloud generation.
More implementation details can be found in released code \url{https://github.com/WoodwindHu/RangeLDM}. 
\section{Experiments}
\label{sec:exp}
\subsection{Experimental Setup}
\noindent\textbf{Datasets.}
We evaluate our model on two challenging datasets, KITTI-360 and nuScenes. KITTI-360 has 81,106 LiDAR readings from 9 sequences in Germany, covering diverse scenes. 
We used the first two sequences for validation and trained on the rest. 
NuScenes is a public dataset with 297,737 LiDAR sweeps in the training set and 52,423 in the testing set, collected in Boston and Singapore.

\noindent\textbf{Evaluation metrics.}
Following \cite{zyrianov2022learning}, , we employed three metrics to perform quantitative analysis: Maximum Mean Discrepancy (MMD), Jensen-Shannon divergence (JSD), and Frechet Range Distance (FRD score).
We use a $100\times 100$ 2D histogram on the BEV plane to calculate MMD and JSD metrics.
The Frechet Range Distance (FRD) score~\cite{zyrianov2022learning} is a metric used to evaluate the quality of samples acquired by a generative model, inspired by the FID score for images \cite{FID}. 
To compute the FRD score, we use RangeNet++ \cite{milioto2019rangenet++}, an encoder-decoder-based network for segmentation, which is pre-trained on KITTI-360.

\noindent\textbf{Baselines.}
We evaluated our approach for LiDAR point cloud generation against several competitive methods,
including LiDAR VAE~\cite{caccia2019deep}, LiDAR GAN~\cite{caccia2019deep}, Projected GAN~\cite{sauer2021projected}, LiDARGen~\cite{zyrianov2022learning} and UltraLiDAR~\cite{xiong2023learning}.

\subsection{Unconditional Generation}


\begin{figure}[t]
    \centering
    \includegraphics[width=1\linewidth]{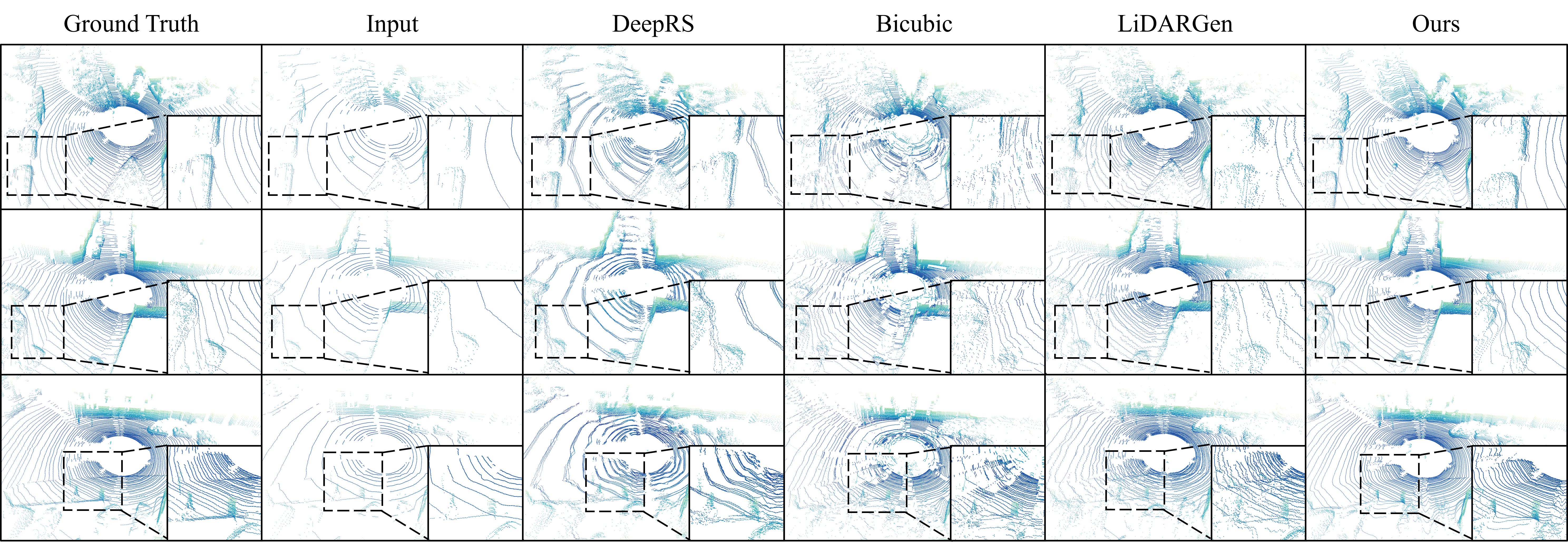}
    \caption{Comparison of upsampling results on KITTI-360. We downsampled the ground truth by a factor of four as the input, and demonstrated the results of different methods on $4\times$-upsampling of the input.}
    \label{fig:upsampling}
\end{figure}

\begin{table}[h]
\centering
\resizebox{0.7\linewidth}{!}{
\begin{tabular}{l|l|ccc}
\hline
Method & Years & $\text{MMD}_\text{BEV}$ $\downarrow$ & $\text{FRD}$ $\downarrow$ & $\text{JSD}_\text{BEV}$ $\downarrow$ \\
\hline
LiDAR GAN~\cite{caccia2019deep} & IROS 2019 & $3.06 \times 10^{-3}$ & 3003.8 & - \\
LiDAR VAE~\cite{caccia2019deep} & IROS 2019 & $1.00 \times 10^{-3}$ & 2261.5 & 0.161 \\
Projected GAN~\cite{sauer2021projected} & NeurIPS 2021 & $3.47 \times 10^{-4}$ & 2117.2 & 0.085 \\
LiDARGen~\cite{zyrianov2022learning} & ECCV 2022 & $3.87 \times 10^{-4}$ & 2040.1 & 0.067 \\
UltraLiDAR~\cite{xiong2023learning} & CVPR 2023 & $1.96 \times 10^{-4}$ & - & 0.071 \\
\hline
Ours & & $\mathbf{3.07 \times 10^{-5}}$ & $\mathbf{1074.9}$ & $\mathbf{0.045}$ \\ 
\hline
\end{tabular}
}
\caption{Unconditional generation results on KITTI-360~\cite{liao2022kitti}. 
}
\label{tab:unconditional}
\end{table}
\noindent\textbf{Quantitative Results on KITTI-360.}
Table~\ref{tab:unconditional} displays the quantitative results of the proposed RangeLDM and competing algorithms. 
The results clearly illustrate that our method significantly outperforms the baselines across all metrics. 
Notably, our results on the MMD metric are an order of magnitude lower than those of other methods. 
This underscores the remarkable expressive capabilities of our model in the context of LiDAR point cloud generation.

\noindent\textbf{Qualitative Results on KITTI-360.}
Figure~\ref{fig:unconditional_kitti360} displays a set of randomly generated samples from competing algorithms, alongside real point cloud samples extracted from the KITTI-360 dataset for comparison. 
We observe that the Projected GAN~\cite{sauer2021projected} exhibits noticeable artifacts in the distant range. 
LiDARGen~\cite{zyrianov2022learning} effectively captures the general layout, but produces rather noisy outputs and lacks the same degree of straight walls as actual samples. 
UltraLiDAR~\cite{xiong2023learning} generates structured and reasonable scenes, yet it only creates voxelized point clouds (as magnified in the first row) without intensity features. 
In contrast, our model consistently surpasses the baseline models and yields results closely resembling real-world data.
For example, we can generate cars (emphasized by red boxes in the first row of Figure~\ref{fig:unconditional_kitti360}) and road bollards (red boxes in the second row of Figure~\ref{fig:unconditional_kitti360}) on the road.
Additionally, our model generates various scenes such as straightway (the first two rows of Figure~\ref{fig:unconditional_kitti360}) and crossroads (the last row of Figure~\ref{fig:unconditional_kitti360}). 
The supplementary materials provide more visualization results.

\begin{wraptable}{r}{0.5\textwidth}
    \centering
    \resizebox{\linewidth}{!}{
\begin{tabular}{l|c}
\hline
Method & Percent prefer ours \\
\hline
Ours vs. VAE~\cite{caccia2019deep} & $98.8\%$ \\
Ours vs. GAN~\cite{caccia2019deep}  & $98.0\%$ \\
Ours vs. ProjectedGAN~\cite{sauer2021projected} & $93.2\%$ \\
Ours vs. LiDARGen~\cite{zyrianov2022learning} & $90.2\%$ \\
Ours vs. UltraLiDAR~\cite{xiong2023learning} & $86.5\%$ \\
\hline
\end{tabular}
}

    \caption{Human study on KITTI-360.}
    \label{tab:human}
\end{wraptable}
\noindent\textbf{Human study on KITTI-360.} 
To assess the perceptual quality, we conducted an A/B test involving a group of 14 researchers with LiDAR expertise.
Following the same evaluation system as~\cite{zyrianov2022learning} and~\cite{xiong2023learning}, we present pairs of randomly selected images from two point clouds and ask participants to determine which one appeared more realistic. 
The results, as displayed in Table~\ref{tab:human}, unequivocally demonstrate the superior visual quality of generation results by our model. 
In most of the cases, testers favored our results over the baselines.

\begin{figure}[t]
    \centering
    \includegraphics[width=0.7\textwidth]{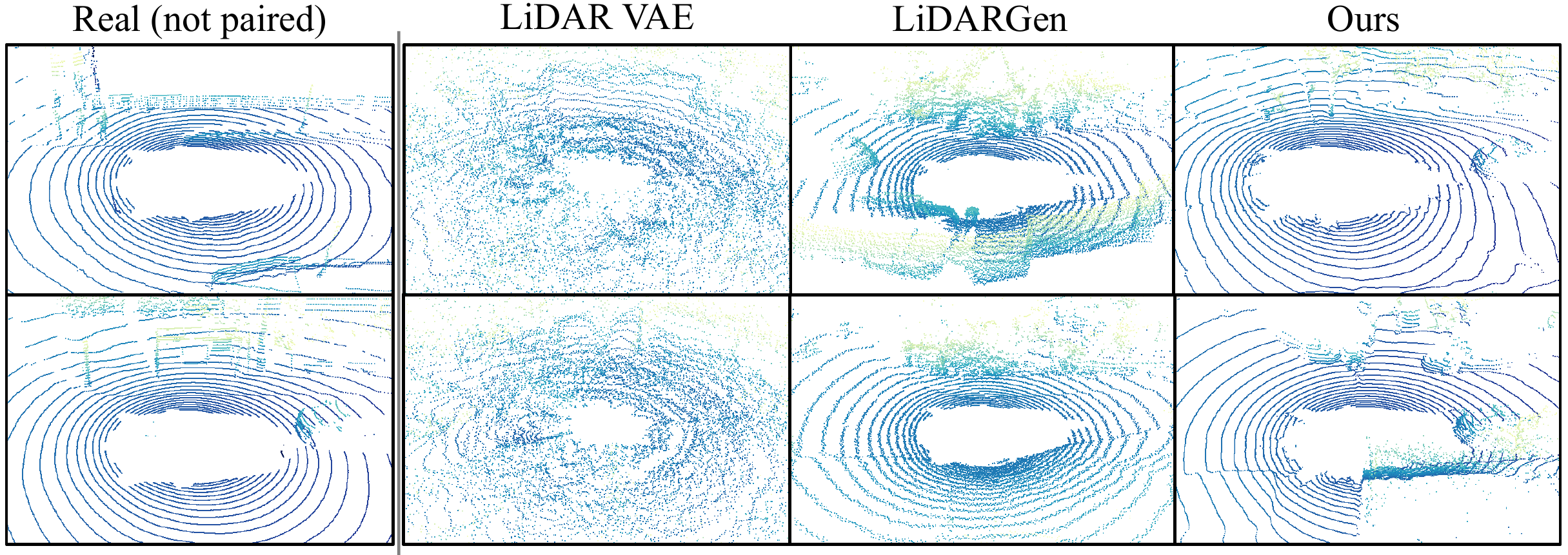} 
    \caption{Qualitative results comparing against LiDAR VAE and LiDARGen for unconditional LiDAR generation on nuScenes. Real point clouds are only for reference. }
    \label{fig:nuscenes}
\end{figure}

\begin{wraptable}{r}{0.4\textwidth}
    \centering
    \resizebox{\linewidth}{!}{
\begin{tabular}{l|cc}
\hline
Method & $\text{MMD}_\text{BEV}$ $\downarrow$ & $\text{JSD}_\text{BEV}$ $\downarrow$  \\
\hline
LiDAR VAE~\cite{caccia2019deep} & $1.1 \times 10^{-3}$ & -\\
$\text{LiDARGen}^\dag$~\cite{zyrianov2022learning} & $1.9 \times 10^{-3}$ & $0.160$ \\
\hline
Ours & $\mathbf{1.9 \times 10^{-4}}$ & $\mathbf{0.054}$ \\
\hline
\end{tabular}
}
\caption{Unconditional generation results on nuScenes dataset~\cite{caesar2020nuscenes}.  \dag: Reproduced by us. }
\label{tab:nuscenes}

\end{wraptable}
\noindent\textbf{Evaluation on the NuScenes Dataset.}
The results listed in Table~\ref{tab:nuscenes} demonstrate that our model outperforms LiDAR VAE and LiDARGen significantly over nuScenes. 
In particular, our results on the MMD metric are an order of magnitude lower than those of other methods, while we reduce to about $1/3$ of the results of competitive methods in terms of the JSD metric. 
Also, Figure~\ref{fig:nuscenes} illustrates the superiority of our model over LiDAR VAE and LiDARGen in terms of qualitative comparison. 
 


\noindent\textbf{Model Size.} 
The number of parameters in LiDARGen~\cite{zyrianov2022learning} and UltraLiDAR~\cite{xiong2023learning} are 29.7M and 40.3M, respectively. 
To ensure fair competition, we limited the capacity of our model to be similar to other approaches. 
Our model comprises 41.4M parameters, consisting of 12.7M parameters for VAE and 28.7M parameters for LDM. 
As illustrated in Table~\ref{tab:unconditional}, our model outperforms the baselines by a large margin under similar model sizes. 


\begin{wraptable}{r}{0.35\textwidth}
    \centering
    \resizebox{\linewidth}{!}{
\begin{tabular}{l|c}
\hline
Method & \makecell{Throughput $\uparrow$ \\ (samples/s)} \\
\hline
LiDARGen~\cite{zyrianov2022learning} & 0.02 \\
$\text{UltraLiDAR}$~\cite{xiong2023learning} & 0.16 \\
\hline
Ours &  \textbf{4.86} \\
\hline
\end{tabular}
}

    \caption{Inference speed on a single RTX 3090 GPU. 
    %
    }
    \label{tab:speed}
\end{wraptable}
\noindent\textbf{Generation Efficiency.}
The proposed dimensionality reduction offers RangeLDM a considerable advantage in terms of generation speed. 
As indicated in Table~\ref{tab:speed}, the proposed method significantly outperforms LiDARGen~\cite{zyrianov2022learning} and UltraLiDAR~\cite{xiong2023learning} in terms of generation speed.
In particular, our model is 200 times faster than LiDARGen and 30 times faster than UltraLiDAR.

\subsection{Conditional Generation}
To evaluate the conditional generation performance of RangeLDM, we conducted experiments on the KITTI-360 dataset over two tasks: LiDAR point cloud upsampling and inpainting.

\begin{table}[t]
\centering
\resizebox{0.90\linewidth}{!}{
\begin{tabular}{c|cccc|ccc}
\hline
& \makecell{Hough \\ Voting} & \makecell{Range-Guided \\ Discriminator} & \makecell{Circular \\ Convolution} & \makecell{Direction \\ Conditioned}  & $\text{MMD}_\text{BEV}$ $\downarrow$ & $\text{FRD}$ $\downarrow$ & $\text{JSD}_\text{BEV}$ $\downarrow$ \\
\hline
(a) & $\times$ & $\times$ & $\times$ & $\times$       & $3.95 \times 10^{-4}$  & $1536.7$	& $0.067$ \\ 
(b) & $\checkmark$ & $\times$ & $\times$ & $\times$       & $6.57\times 10^{-5}$  & 1229.3	&  0.056  \\ 
(c) & $\checkmark$ & $\checkmark$ & $\times$ & $\times$       & $4.72\times 10^{-5}$  & $1103.1$	& $0.051$   \\ 
(d) & $\checkmark$   & $\checkmark$ & $\checkmark$   & $\times$       &  $3.90 \times 10^{-4}$ & 1797.2	& 0.078  \\ 
(e) & $\checkmark$ & $\checkmark$ & $\checkmark$   & $\checkmark$    & $\mathbf{3.07 \times 10^{-5}}$ & $\mathbf{1074.9}$ & $\mathbf{0.045}$  \\ 
\hline
\end{tabular}
} 
\caption{Main ablation study. }
\label{tab:ablation}
\end{table}

\noindent\textbf{LiDAR Point Cloud Upsampling.}
We obtain sparse input point clouds by selecting a subset of 16 beams from the raw 64-beam sensors, in line with LiDARGen. 
We compared our approach against PUNet~\cite{yu2018pu}, DeepRS~\cite{chen2022deep}, Grad-PU~\cite{Grad-PU}, bicubic interpolation, Nearest Neighbor (NN) interpolation and LiDARGen. 

\begin{wraptable}{r}{0.5\textwidth}
    \centering
    \resizebox{\linewidth}{!}{
\begin{tabular}{l|c|ccc}
\hline
Method & Years & $\text{MAE}$ $\downarrow$ & $\text{Accuracy}$ $\uparrow$ & $\text{IoU}$ $\uparrow$\\
\hline
PUNet~\cite{yu2018pu} & 2018 & 6.88 & - & - \\
DeepRS~\cite{chen2022deep} & 2022 & 3.96 & - & - \\
Grad-PU~\cite{Grad-PU} & 2023 & 5.09 & - & - \\
\hline
Bicubic & -& 2.60 & 0.265 & 0.166 \\
NN & - & 2.18 & 0.546 & 0.394 \\
LiDARGen~\cite{zyrianov2022learning} & 2022 & 1.23 & 0.608 & 0.449 \\
\hline
Ours & & $\mathbf{0.89}$ & $\mathbf{0.722}$ & $\mathbf{0.566}$ \\ 
\hline
\end{tabular}
}
\caption{LiDAR upsampling. 
}
\label{tab:upsample}

\end{wraptable}
Quantitatively, in addition to measuring Mean Absolute Error (MAE) in the range view, we also employed RangeNet++ semantic segmentation to evaluate the quality of upsampled results. 
As shown in Table~\ref{tab:upsample}, the proposed method outperforms competing methods in all metrics, including MAE, per-point segmentation accuracy, and segmentation Intersection over Union (IoU). 

Qualitatively, the visualization comparison with several competitive methods presented in Figure~\ref{fig:upsampling} indicates that the proposed upsampling method exhibits results highly similar to the ground truth data, showing clear object details and consistent LiDAR scan lines. 
We provide additional visualization results in the supplementary materials.

\begin{figure}[t]
    \centering
    \includegraphics[width=0.7\textwidth]{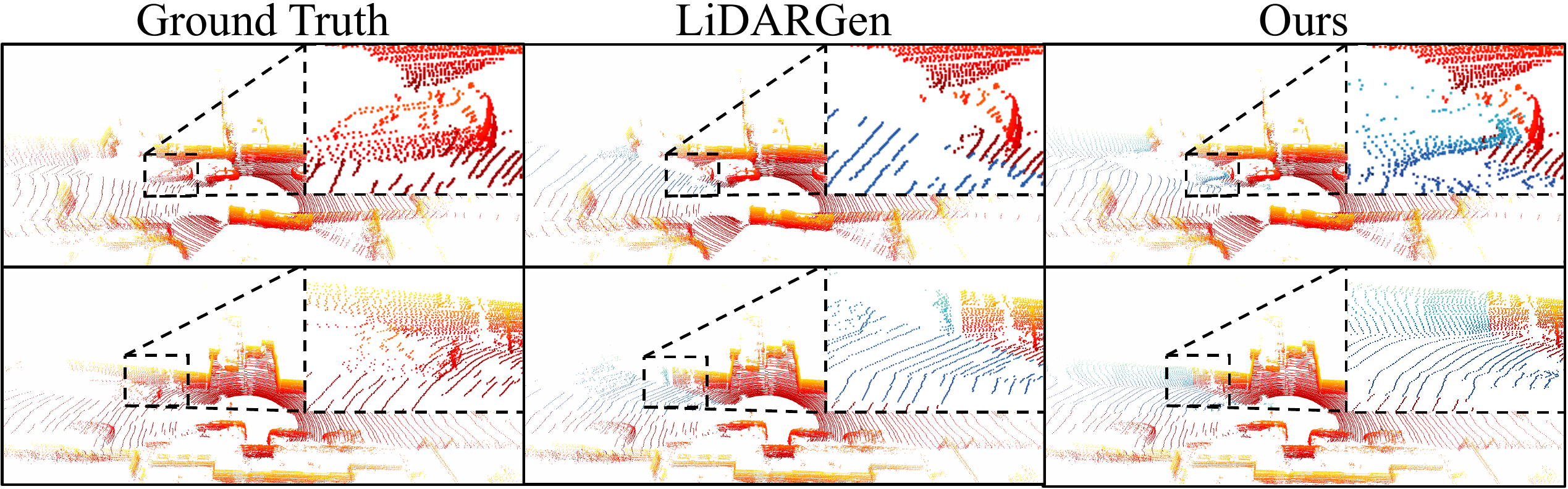} 
    \caption{Inpainting results. Left: ground truth point clouds. Middle and right (red): input point clouds. Middle and right (blue): recovered point clouds. }
    \label{fig:inpainting}
\end{figure}
\noindent\textbf{LiDAR Point Cloud Inpainting.}
To mimic the case of missing data in LiDAR point clouds, we mask point clouds in front of the vehicle within a $22.5^\circ$ range and perform inpainting using LiDARGen and our method. 
Our results achieve much less MAE ($0.190$) between the reconstructed results and ground truth compared to that of LiDARGen ($0.367$). 
As shown in Figure~\ref{fig:inpainting}, LiDARGen generates roads but fails to recover the masked car (the first row) and the wall (the second row), while we are able to reconstruct the details. 
More results are presented in the supplementary materials.

\subsection{Ablation Study}
\label{subsec:ablation}
All ablation studies are conducted on the KITTI-360 dataset. 

\begin{figure}[t]
    \centering
    \includegraphics[width=1.0\linewidth]{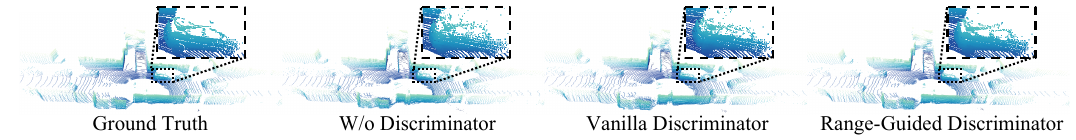}
    \caption{VAE reconstruction results. The proposed VAE with a range-guided discriminator reconstructs point clouds with less noise and more precise object structures.}
    \label{fig:discriminator}
\end{figure}
\begin{table}[t]
\centering
\resizebox{0.8\linewidth}{!}{
\begin{tabular}{c|c|c|cccc}
\hline
 Model & VAE & \makecell{Model \\ Size (M)} & $\text{MMD}_\text{BEV}$ $\downarrow$ & $\text{FRD}$ $\downarrow$ & $\text{JSD}_\text{BEV}$ $\downarrow$ & \makecell{Throughput $\uparrow$ \\ (samples/s)} \\
\hline
 Ours-DMS & $\times$ & 40.3 & $1.03\times 10^{-4}$ & $1392.2$ & $0.062$ & 1.52 \\ 
 Ours-DM & $\times$ & 114.7  & $4.14 \times 10^{-5}$  & $\mathbf{899.0}$ & $\mathbf{0.040}$ & 0.64 \\ 
\hline
 Ours & $\checkmark$ & 41.4 & $\mathbf{3.07 \times 10^{-5}}$ & 1074.9 & 0.045 & $\mathbf{4.86}$ \\ 
\hline
\end{tabular}
} 
\caption{Ablation of latent diffusion. Model without latent diffusion requires more parameters to achieve similar performance with the latent diffusion model. We generated 1000 samples for calculating the throughput.}
\label{tab:DM}
\end{table}

\noindent\textbf{Main Ablation.}
As demonstrated in Table~\ref{tab:ablation}, we investigate the contribution of each component. 
Starting from the backbone model (a) without any component, we gradually add each component for evaluation. 
By applying Hough Voting to the training, the performance of variant (b) improves significantly, demonstrating the huge impact of the correct range-view data distribution. 
With the addition of the range-guided discriminator component, model (c) achieves better performance, thus clarifying the effectiveness of the geometry-sensitive module. 
Compared to model (c), the performance of model (d) decreases because circular convolution generates LiDAR scenes with {\it arbitrary} directions due to rotation invariance. 
However, when combined with direction-conditioned generation that guides the model to generate point clouds in the correct direction, the overall performance improves, as demonstrated by model (e).
We refer the readers to the supplementary materials for qualitative results.

\noindent\textbf{Latent diffusion.}
We explore the contribution of latent diffusion to the generation quality and generation speed of the model.
As shown in Table~\ref{tab:DM}, we constructed two variants of different sizes that directly generate range images with diffusion models, denoted as ``Ours-DMS'' (with a smaller size) and ``Ours-DM'' (with a larger size), for comparison. 
It turns out that a much larger model size is required for ``Ours-DM'' to achieve competitive performance with ``Ours'', due to the multitude of high-frequency details contained in the range image. 
Additionally, the superior throughput of ``Ours'' demonstrates the efficiency of latent diffusion.

\begin{wraptable}{r}{0.55\textwidth}
    \centering
\centering
\resizebox{\linewidth}{!}{
\begin{tabular}{c|cccc}
\hline
Discriminator & $\text{PSNR}_\text{range}$ $\uparrow$ & $\text{MAE}$ $\downarrow$ & $\text{FRD}$ $\downarrow$ & $\text{CD}$ $\downarrow$  \\
\hline
No   & 26.77 & 0.0195 & 532.9 & 0.0808 \\
Vanilla &  26.70 & 0.0189 & 496.7 & 0.0726 \\
Range-Guided   & $\mathbf{27.19}$ & $\mathbf{0.0186}$ & $\mathbf{483.6}$ & $\mathbf{0.0676}$ \\
\hline
\end{tabular}
}
\caption{Ablation on the VAE architecture. }
\label{tab:range-emphasised}
\end{wraptable}
\noindent\textbf{VAE Architecture.}
We investigate the contribution of the proposed range-guided discriminator by comparing our VAE with two baselines: 1) VAE without discriminator; and 2) VAE with vanilla discriminator ({\it i.e.}, using the original 2D convolution in the discriminator).
We evaluate the reconstruction performance of VAEs with 2D and 3D metrics, including $\text{PSNR}$,  $\text{MAE}$,  $\text{FRD}$ and Chamfer Distance (CD) \cite{fan2017point}. 
As listed in Table~\ref{tab:range-emphasised}, the VAE with the proposed range-guided discriminator outperforms the baselines in both 2D and 3D reconstruction metrics. 
This gives credits to the ability of perceiving local 3D structures by the range-guided discriminator. 
The qualitative results presented in Figure~\ref{fig:discriminator} also indicate that the proposed range-guided discriminator ensures the point clouds are reconstructed with less noise and more precise object structures.

\section{Conclusions}
\label{sec:conc}

We propose a novel RangeLDM model to generate realistic range-view LiDAR point clouds at a fast speed.
We ensure the quality of projection from point clouds to range images with correct distribution via Hough Voting. 
Then we compress range images to latent features with a VAE, and train a diffusion model in the lower-dimensional latent space. 
Additionally, we enhance the range-image reconstruction quality of the VAE with a range-guided discriminator. 
Experiments conducted on KITTI-360 and nuScenes datasets demonstrate the superior generation quality and sampling efficiency of our method. 
In future, we will explore generating labeled point clouds and creating corner-case data, such as in car accidents and extreme weather conditions, for potential applications in robust self-driving.

%
%
\bibliographystyle{splncs04}
\bibliography{main}
\clearpage
\end{document}